\documentclass[journal]{IEEEtran}

\usepackage[T1]{fontenc}
\usepackage[utf8]{inputenc}
\usepackage{amsmath,amssymb,amsfonts}
\usepackage{graphicx}
\usepackage{booktabs}
\usepackage{array}
\usepackage{multirow}
\usepackage[table]{xcolor}
\usepackage{algorithm}
\usepackage{algpseudocode}
\usepackage{hyperref}
\usepackage{microtype}
\usepackage{cite}
\usepackage{balance}
\usepackage{dblfloatfix}   

\hypersetup{
  colorlinks=true,
  linkcolor=blue!60!black,
  citecolor=blue!60!black,
  urlcolor=blue!60!black
}

\newcommand{\shat}{\hat{s}}

\newcommand{\scoreS}{\mathrm{score}_{\text{SWAM}}}

\title{SWAM: Saliency-Weighted Asymmetric Merging for Efficient Vision Transformers}

\author{Utkarsh Kumar Singh, Manish Das, \IEEEmembership{Student Member,~IEEE}, Shiv Ram Dubey, \IEEEmembership{Senior Member,~IEEE},
and Satish Kumar Singh, \IEEEmembership{Senior Member, ~IEEE}
\thanks{U.K. Singh, M. Das, S.R. Dubey and S.K. Singh are with the Computer Vision and Biometrics Lab, Department of Information Technology, Indian Institute of Information Technology Allahabad, Prayagraj, Uttar Pradesh-211015, India (e-mail: mml2024007@iiita.ac.in, pmm2023001@iiita.ac.in, srdubey@iiita.ac.in, sk.singh@iiita.ac.in).}
}

\begin{document}

\maketitle

\begin{abstract}
Vision Transformers (ViTs) achieve strong performance on image recognition tasks, yet their quadratic self-attention complexity over input tokens creates prohibitive computational costs for practical deployment.  Token reduction methods offer a practical solution by progressively reducing the token sequence during the forward pass.  We propose \textbf{SWAM} (Saliency-Weighted Asymmetric Merging), a training-based token merging framework that extends the Bipartite Soft Matching (BSM)  with two saliency-aware mechanisms. 
We introduce a two-step matching formulation in which cosine similarity decides candidate pairs, while a multiplicative maximum saliency penalty regulates their selection for merging. We then further use asymmetric exponential saliency-weighted aggregation, which allows more salient tokens to contribute more strongly to the merged representation.
A lightweight MLP-based saliency prediction module ($ 75\text{k}$ parameters per stage, $1\%$ of backbone parameters) supervised via attention rollout is trained end-to-end jointly with a DeiT-Small backbone using a composite cross-entropy, knowledge-distillation, and saliency-alignment objective.  On CIFAR-100, TinyImageNet, and ImageNet-100, under a conservative schedule that reduces tokens from $196{\to}100$ patch tokens, with the CLS token retained throughout, 27.5\% FLOPs reduction and $1.35{\times}$ speedup, SWAM matches or exceeds baseline across the reported results.  Under an aggressive schedule ($196{\to}70$ tokens with the CLS token kept throughout, 33.8\% FLOPs reduction, and $1.46{\times}$ speedup), accuracy gains are achieved on ImageNet-100 ($+0.23\%$) with minimal losses on CIFAR-100 and TinyImageNet. 

\end{abstract}

\begin{IEEEkeywords}
vision transformer, token merging, token reduction, saliency, efficient inference, DeiT
\end{IEEEkeywords}

\section{Introduction}\label{sec:intro}

The Vision Transformer (ViT) \cite{dosovitskiy2021vit} has become a dominant architecture for image recognition, processing images as sequences of patch tokens and passing them through stacked multi-head self-attention (MHSA) layers.  Models such as DeiT-Small achieve strong ImageNet-scale performance while retaining the standard Transformer architecture.  However, self-attention requires $\mathcal{O}(N^{2}{\cdot}d)$ computation per layer, where $N$ is the
number of tokens and $d$ the embedding dimension.  For DeiT-Small at
$224{\times}224$ resolution ($N{=}196$ patch tokens $+$ 1 CLS token, $d{=}384$,
12 blocks), this amounts to approximately $4.25$\,GFLOPs per forward pass. This computational requirement can become restrictive when vision transformers are deployed under limited latency or computational budgets.

Token reduction methods address this by progressively removing tokens during the
forward pass, reducing attention cost quadratically in the number of removed
tokens.   Among token-merging approaches, Bipartite Soft Matching (BSM), used in ToMe \cite{bolya2023tome}, identifies candidate token pairs according to feature similarity and merges the most similar pairs. While similarity provides a useful measure of redundancy, it does not explicitly indicate how important an individual token is for the classification task.
This leads to two practical limitations. First, a highly informative token may still participate in a merge when it is strongly similar to another token, because similarity-based pair selection does not explicitly account for token importance. Second, once a pair is selected, conventional size-weighted averaging treats the merged representations according to their accumulated sizes rather than their semantic importance. Consequently, informative token content may be weakened during repeated token reduction.

To adress this, we propose \textbf{SWAM} (Saliency-Weighted Asymmetric
Merging), a training-based token merging framework that introduces a specific saliency-aware formulation for pair ranking and feature aggregation. Existing importance-aware reduction methods have demonstrated the usefulness of token relevance for guiding selection and fusion decisions \cite{liang2022evit,wei2023tps,lee2024mctf}.
Rather than treating the incorporation of importance into merging as a new principle,
SWAM focuses on how saliency influences the two operations. Candidate destinations
are first identified through cosine similarity, after which a multiplicative penalty based on the maximum predicted saliency of each pair determines its merge ranking. This formulation reduces the preference for highly salient pairs when their feature similarity is positive. For selected pairs, SWAM uses exponential saliency-dependent weights to construct
the merged representation, assigning greater influence to more salient tokens without
using accumulated token size as an aggregation weight. A lightweight saliency prediction module provides the importance estimates, with attention-rollout supervision \cite{abnar2020rollout} applied at the first reduction stage. The resulting framework therefore differs in its pair-ranking criterion, feature aggregation rule, and learned saliency estimation, rather than in the general use of token importance during merging.
The main contributions of our work are : 
\begin{itemize}
  \item \textbf{Multiplicative max-saliency pair scoring.} We introduce a two-step formulation that identifies token pairs through cosine similarity and ranks candidate merges using exponential penalty based on their maximum predicted saliency.

  \item \textbf{Asymmetric saliency-weighted merging.}  A combination step that
    weights each token's contribution by an exponential function of its
    predicted saliency, preserving the content of the more important token.

    \item \textbf{Lightweight saliency prediction.}
    Three two-layer MLP predictors estimate token importance at the reduction stages. The first-stage predictor receives explicit attention-rollout supervision, while subsequent predictors are optimized through downstream task gradients propagated by differentiable feature aggregation, adding approximately $1\%$ parameter overhead.

  \item \textbf{Evaluation under progressive token compression.}
    We evaluate SWAM on CIFAR-100, TinyImageNet, and ImageNet-100 using conservative and aggressive token-reduction schedules and report  classification accuracy with FLOPs, throughput, latency, and
    memory consumption.
\end{itemize}

\section{Related Work}\label{sec:related}

The ViT architecture~\cite{dosovitskiy2021vit} divides an image into $N$
patch tokens, prepends a learnable CLS token, and processes the sequence through
$L$ transformer blocks applying MHSA followed by  FFN with residual connections.  DeiT~\cite{touvron2021deit} demonstrates that strong ImageNet performance is achievable without large-scale private pretraining via knowledge distillation. To improve the computational efficiency of Vision Transformers, several token-reduction strategies have been proposed, including token pruning, token merging, and hybrid approaches.

\subsection{Token Pruning}
Token pruning improves Vision Transformer efficiency by removing tokens that are considered less important or informative during inference. DynamicViT~\cite{rao2021dynamicvit} learns binary keep/drop decisions at multiple stages.  EViT~\cite{liang2022evit} fuses inattentive tokens into a single attentive summary token using CLS attention scores. A-ViT~\cite{yin2022avit} introduces an adaptive token-halting mechanism that learns a halting probability for each spatial token. IA-RED$^2$ \cite{iared} introduces modules that are interpretable and allow identifying and progressively removing redundant patches at different layers. Several approaches further make token identification dependent on the input or the target computational budget. Evo-ViT \cite{xu2022evo} uses class attention to differentiate informative and less informative tokens and update them through different computational paths. ATS \cite{ats} introduces a parameter-free approach that scores and adaptively samples significant tokens, allowing the preserved token count to vary across input images. SPViT \cite{spvit} employs an attention-based token selector together with soft pruning, where less informative tokens are combined into a package token rather than being completely discarded. Patch slimming \cite{patch_slimming} follows a top-down approach that identifies important tokens in deeper layers to guide patch selection in the earlier layers. TokenLearner \cite{ryoo2021tokenlearner} reduces tokens by learning a compact set of adaptive visual tokens from the input representation. These approaches reduce computation effectively, but pruning may discard token information completely once a token is removed. Token merging provides an alternative in which redundant tokens are combined rather then eliminated, allowing some part of their information to remain in the representation. 

\subsection{Token Merging and Hybrid Methods}

Token Merging (ToMe) \cite{bolya2023tome} introduced Bipartite Soft Matching (BSM), where tokens are divided into two sets and candidate pairs are identified according to feature similarity. Several later approaches extend token reduction by incorporating additional signals. TPS \cite{wei2023tps} transfers information from pruned tokens to retained representation. MCTF \cite{lee2024mctf} combines multiple criteria, including token similarity, attention information, and accumulated token size, to improve merging decisions. DeepCompressViT \cite{deepcompress}, and TinyNext-ViT \cite{tinynext} further improve token efficiency through compact token modeling,  while related methods further improve token interaction and fusion for efficient representation learning \cite{b16,b17}. LTMP \cite{bonnaerens2023ltmp} learns threshold-based decisions for dynamically merging or pruning tokens. Compared with these methods, SWAM differs in formulation where cosine similarity first determines candidate pairs, then a multiplicative maximum saliency penalty for merging ranking. 

\subsection{Attention Rollout}

Attention information has been used as an indicator of token importance in Vision Transformers. Attention rollout \cite{abnar2020rollout} aggregates attention across multiple Transformer layers to estimate the cumulative information flow from the CLS token to image patches. Such information provides a useful supervisory signal because it captures token relevance over several attention layers rather than relying on a single attention map. SWAM uses attention rollout as supervision for the saliency prediction module at the first reduction stage.

\section{Methodology}\label{sec:method}

\begin{figure*}[!t]
  \centering
  \includegraphics[width=\textwidth]{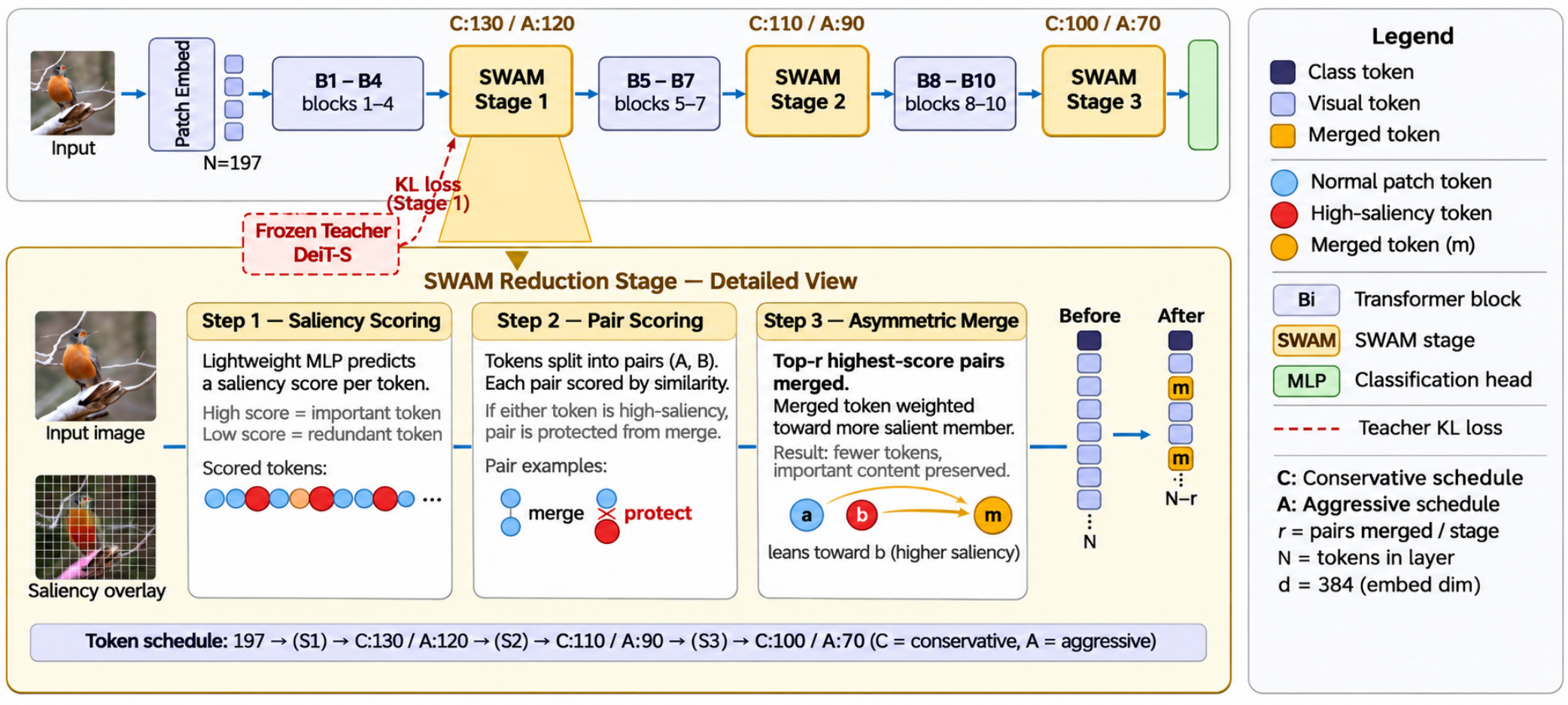}
  \caption{%
    \textbf{SWAM pipeline.}  \emph{Top}: DeiT-Small backbone with three SWAM
    reduction stages inserted after blocks 4, 7, and 10.  Patch-token counts decrease
    from 196 to 100 (conservative, C) or 70 (aggressive, A) while the CLS token is retained throughout all reduction stages.
    \emph{Bottom}: Detail of a single stage.  The saliency head predicts
    per-token importance scores that feed into multiplicative max-saliency pair
    scoring (Eq. \ref{eq:score}) and asymmetric saliency-weighted merging
    (Eq. \ref{eq:merge}).
  }
  \label{fig:pipeline}
\end{figure*}

\subsection{Overview}

SWAM is a training-based token reduction framework built on DeiT-Small. Three progressive reduction stages are introduced after Transformer blocks 4, 7, and 10. Under the conservative configuration, the patch-token count is progressively reduced as $196{\to}130{\to}110{\to}100$, while the CLS token is retained throughout all reduction stages. Therefore, the corresponding total sequence lengths, including the CLS token, are $197{\to}131{\to}111{\to}101$. Under the aggressive configuration, the patch-token count is reduced from 196 to 70, corresponding to a final sequence length of 71 including the CLS token. At each stage, a lightweight saliency
head estimates a scalar importance score for every patch token, which modulates (i) pair selection, biasing it away from important tokens, and (ii) the merged representation, biasing it toward more important tokens. The full pipeline is illustrated in Fig. \ref{fig:pipeline}.

\subsection{Bipartite Soft Matching Foundation}

For $N$ tokens at a reduction stage where $r$ tokens are to be removed, BSM
partitions tokens into disjoint sets $A$ (odd-indexed) and $B$ (even-indexed),
finds each $a_i\!\in\!A$'s best-matching $b_{j^*}$ by cosine similarity, selects
the top-$r$ highest-similarity pairs, and merges each by size-weighted average:
\begin{equation}
  m = \frac{s_a\,a + s_b\,b}{s_a + s_b},
  \label{eq:tome_merge}
\end{equation}
where $s_a, s_b$ are accumulated size counters.  SWAM replaces the selection
and combination steps with the two novel components.

\subsection{Saliency Prediction Module}

At each reduction stage a two-layer MLP predicts a scalar saliency score for
each token $z_i$: $s_i = W_2\,\mathrm{GELU}(W_1 z_i + b_1) + b_2, \label{eq:mlp}$ with $W_1\!\in\!\mathbb{R}^{192\times384}$ and $W_2\!\in\!\mathbb{R}^{1\times192}$. The raw scalar output $\shat_i = s_i$ is used directly as the token's saliency score, without additional bounding.  This yields approximately $74\text{k}$ parameters per stage (${\approx}222\text{k}$ total), corresponding to about $1\%$ of the DeiT-Small backbone parameters.

\subsection{Saliency Supervision via Attention Rollout}

A trained full-token DeiT-S baseline is used to initialize the teacher model.
The teacher is subsequently frozen, and its attention maps are captured to
construct the saliency-supervision target. For each Transformer block, the
attention matrices are first averaged across the attention heads:
\begin{equation}
  \bar{\mathbf{A}}_{\ell}
  =
  \frac{1}{H}
  \sum_{h=1}^{H}
  \mathbf{A}_{\ell,h},
  \label{eq:headavg}
\end{equation}
where $H$ denotes the number of attention heads. Residual attention is then
incorporated and row-normalized as
\begin{equation}
  \widetilde{\mathbf{A}}_{\ell}
  =
  \operatorname{RowNorm}
  \left(
    \frac{\bar{\mathbf{A}}_{\ell}+\mathbf{I}}{2}
  \right).
  \label{eq:rollout_matrix}
\end{equation}
Since the first SWAM reduction stage is inserted after Transformer block~4,
the corresponding cumulative attention rollout is $
  \mathbf{R}^{(4)}
  =
  \widetilde{\mathbf{A}}_{4}
  \widetilde{\mathbf{A}}_{3}
  \widetilde{\mathbf{A}}_{2}
  \widetilde{\mathbf{A}}_{1}.
  \label{eq:rollout}
$
The CLS-to-patch entries of $\mathbf{R}^{(4)}$ are normalized to obtain the
196-dimensional target distribution $\mathbf{t}$. Explicit rollout supervision
is applied only at the first reduction stage because its input tokens still
maintain a one-to-one correspondence with the original image patches. At later
stages, the patch tokens have already undergone merging.

The saliency loss is defined as
\begin{equation}
  \mathcal{L}_{\mathrm{sal}}
  =
  D_{\mathrm{KL}}
  \left(
      \mathbf{t}
      \;\middle\|\;
      \operatorname{softmax}
      \left(\hat{\mathbf{s}}^{(1)}\right)
  \right).
  \label{eq:lsal}
\end{equation}

 At further stages, saliency scores are learned through classification and knowledge distillation losses. Gradients propagate through differentiable weights used in feature aggregation (Eq. \ref{eq:merge}), while the discrete operations do not provide gradients through their ranking decisions. Consequently, later layers scores represent task-adaptive importance estimates learned through their influence on merged features, rather than directly supervised saliency targets.

\subsection{Component 1: Multiplicative Max-Saliency Pair Scoring}

As an additive comparison, candidate pairs can be scored using their mean predicted saliency:
\begin{equation}
  \mathrm{score}_{\mathrm{add}}(a,b)
  =
  \mathrm{sim}(a,b)
  -
  \alpha\cdot
  \frac{\hat{s}_a+\hat{s}_b}{2}.
  \label{eq:additive}
\end{equation}

For a pair containing one highly salient token and one less salient token,
the mean saliency may provide only a moderate penalty. SWAM instead modulates
the pair similarity using the maximum predicted saliency:
\begin{equation}
  \scoreS(a,b)
  =
  \mathrm{sim}(a,b)\cdot
  \exp\!\left(
    -\alpha\max(\hat{s}_a,\hat{s}_b)
  \right),
  \label{eq:score}
\end{equation}

For positively correlated candidate pairs, increasing the saliency of either token reduces the multiplicative factor and consequently lowers the merge score. Thus, among highly similar candidate pairs, pairs containing high saliency  tokens are less likely to be selected, whereas pairs that are
both similar and relatively low-saliency tend to receive larger merge scores.
The $r$ source tokens with the highest resulting scores are selected for
merging with their corresponding best-matching destination tokens.

\subsection{Component 2: Asymmetric Saliency-Weighted Merging}

Even when token merging is unavoidable, conventional size-weighted averaging
can dilute the representation of a more informative token. SWAM instead uses
the predicted saliency scores to control the contribution of each token to the
merged representation. Let $\mathcal{I}_j$ denote the set of selected source tokens in $A$ whose
best-matching destination is token $b_j\in B$:
\begin{equation}
  \mathcal{I}_j
  =
  \left\{
    i \mid a_i \text{ is selected for merging into } b_j
  \right\}.
\end{equation}

Each token is assigned an exponential saliency-dependent weight,
\begin{equation}
  w_{a_i}=e^{\beta\hat{s}_{a_i}},
  \qquad
  w_{b_j}=e^{\beta\hat{s}_{b_j}},
\end{equation}
where $\beta$ controls the strength of the asymmetric weighting. The updated
representation of destination token $b_j$ is then computed as
\begin{equation}
  \mathbf{m}_j
  =
  \frac{
    w_{b_j}\mathbf{b}_j
    +
    \displaystyle\sum_{i\in\mathcal{I}_j}
    w_{a_i}\mathbf{a}_i
  }{
    w_{b_j}
    +
    \displaystyle\sum_{i\in\mathcal{I}_j}
    w_{a_i}
  },
  \label{eq:merge}
\end{equation}

Consequently, tokens with larger predicted saliency contribute more strongly
to the merged representation, allowing informative semantic content to be
preserved during token reduction. The accumulated token size is updated
separately by summing the sizes of all source tokens assigned to the same
destination token.


\begin{algorithm}[!t]
\caption{SWAM StageReducer Forward Pass}
\label{alg:swam}
\begin{algorithmic}[1]
\Require Patch-token matrix
         $Z\in\mathbb{R}^{N\times d}$,
         token-size vector
         $\mathbf{s}\in\mathbb{R}^{N}$,
         target token count $N_{\mathrm{keep}}$,
         hyperparameters $\alpha,\beta$
\Ensure Reduced token matrix
        $Z'\in\mathbb{R}^{N_{\mathrm{keep}}\times d}$
        and updated size vector $\mathbf{s}'$

\State $r \leftarrow N-N_{\mathrm{keep}}$
\State $\hat{\mathbf{s}} \leftarrow \mathrm{MLP}(Z)$
\Comment{predicted saliency}

\State $A \leftarrow \{z_1,z_3,\ldots\}$
\State $B \leftarrow \{z_2,z_4,\ldots\}$

\For{each $a_i\in A$}
    \State
    $j^*(i)
    \leftarrow
    \arg\max_j
    \mathrm{sim}(a_i,b_j)$
    \Comment{best destination}

    \State
    $q_i
    \leftarrow
    \mathrm{sim}(a_i,b_{j^*(i)})
    \exp\!\left[
       -\alpha
       \max
       \left(
          \hat{s}_{a_i},
          \hat{s}_{b_{j^*(i)}}
       \right)
    \right]$
\EndFor

\State
$\mathcal{R}
\leftarrow
\operatorname{TopK}_{\mathrm{largest}}
\left(
\{q_i\},r
\right)$
\Comment{highest-scoring source tokens selected for merging}

\For{each destination token $b_j\in B$}

    \State
    $\mathcal{I}_j
    \leftarrow
    \left\{
       i\in\mathcal{R}
       \mid
       j^*(i)=j
    \right\}$

    \State
    $w_{b_j}
    \leftarrow
    \exp(\beta\hat{s}_{b_j})$

    \For{each $i\in\mathcal{I}_j$}
        \State
        $w_{a_i}
        \leftarrow
        \exp(\beta\hat{s}_{a_i})$
    \EndFor

    \State
    $\displaystyle
    b'_j
    \leftarrow
    \frac{
        w_{b_j}b_j
        +
        \sum_{i\in\mathcal{I}_j}
        w_{a_i}a_i
    }{
        w_{b_j}
        +
        \sum_{i\in\mathcal{I}_j}
        w_{a_i}
    }$

    \State
    $\displaystyle
    s'_j
    \leftarrow
    s_{b_j}
    +
    \sum_{i\in\mathcal{I}_j}s_{a_i}$

\EndFor

\State
$A_{\mathrm{keep}}
\leftarrow
A\setminus\{a_i:i\in\mathcal{R}\}$

\State
$Z'
\leftarrow
[A_{\mathrm{keep}},B']$

\State
$\mathbf{s}'
\leftarrow
[\mathbf{s}_{A_{\mathrm{keep}}},
 \mathbf{s}_{B'}]$

\end{algorithmic}
\end{algorithm}

\subsection{Training Objective}

The SWAM student is optimized using ground-truth classification,
hard-label knowledge distillation, and saliency supervision. The total training objective is
\begin{equation}
  \mathcal{L}_{\mathrm{total}}
  =
  0.5\,\mathcal{L}_{\mathrm{CE}}
  +
  0.5\,\mathcal{L}_{\mathrm{KD}}
  +
  \lambda_e\,\mathcal{L}_{\mathrm{sal}},
  \label{eq:loss}
\end{equation}
where $\mathcal{L}_{\mathrm{CE}}$ is the cross-entropy loss with respect to
the ground-truth label, $\mathcal{L}_{\mathrm{KD}}$ is the cross-entropy loss
with respect to the teacher's hard prediction, and $\lambda_e$ denotes the
epoch-dependent saliency-loss coefficient.


\section{Experimental Setup}\label{sec:setup}

\subsection{Datasets}

\textbf{CIFAR-100}: CIFAR-100 contains 50,000 training images and 10,000 testing images of 100 class categories of original size  $32{\times}32$ and is resized to $224{\times}224$ to match the input resolution of the DeiT-S backbone. Models are trained and evaluated for 15 epochs.  \noindent\textbf{TinyImageNet}:  TinyImageNet consists of 200 object categories with 100,000 training images and 10,000 validation images. The original image size is of $64{\times}64$ resized to $224{\times}224$ for DeiT-S backbone. Depending on the experimental settings are trained and tested for 15 to 20 epochs.  \noindent\textbf{ImageNet-100}: ImageNet-100 contains 130k training and 5k validation images across 100 classes, processed at $224{\times}224$ resolution. Since DeiT-S is pretrained on full ImageNet-1K, results on this subset represent in-domain fine-tuning. Training and testing is performed for 20-25 epochs depending on the experimental configuration.

\subsection{Compression Schedules}
SWAM follows a progressive token compression strategy in which redundant patch tokens are merged at three stages, after Transformer blocks 4, 7, and 10. Rather than removing a large number of tokens at a single stage, the token sequence is reduced gradually so that the network can preserve informative representations across successive layers. The CLS token is retained throughout  all reduction stages, while only the patch tokens are compressed. Two operating points are evaluated to study the trade-off between accuracy and computational efficiency. As summarized in Table \ref{tab:compression}, the conservative schedule reduces the patch-token count from 196 to 100, and the aggressive schedule reduces it to 70. These settings provide different trade-offs between computational efficiency and recognition accuracy.

\begin{table}[!t]
  \centering
  \caption{Compression schedules evaluated.}
  \label{tab:compression}
  \begin{tabular}{lccc}
    \toprule
    Schedule & Patch-token counts & FLOPs $\downarrow$ & Speedup \\
    \midrule
    Conservative & $196{\to}130{\to}110{\to}100$ & 27.5\% & $  1.35{\times}$ \\
    Aggressive   & $196{\to}120{\to}90{\to}70$   & 33.8\% & $  1.46{\times}$ \\

    \bottomrule
  \end{tabular}
\end{table}

\subsection{Baseline, Metrics, and Training Configuration}

The primary reference is a full-token DeiT-S model without token reduction. The baseline is trained using ground-truth cross-entropy, whereas SWAM employs the composite objective in Eq. \ref{eq:loss}, incorporating classification, hard-label knowledge distillation, and saliency supervision. Additionally, the SWAM classification head is initialized from the trained baseline classifier. Therefore, the accuracy comparison evaluates the complete SWAM training and inference framework against the full-token reference, rather than isolating the effect of token merging independently of the additional training objectives. We report top-1 accuracy, throughput (img/s, batch size 128, single GPU), latency (ms/batch), FLOPs, and peak GPU memory.

All experiments are trained using a batch size of 256 with the AdamW optimizer and a weight decay of 0.05. The learning rate follows a cosine-annealing schedule with warm restarts. An exponential moving average (EMA) of the model parameters with a decay factor of 0.9999 is used. The saliency-loss coefficient increases to a maximum value of $\lambda=0.25$. For SWAM, the multiplicative saliency-scoring coefficient is set to $\alpha=1.0$, while the asymmetric merging coefficient is set to $\beta=4.0$. Training is performed using automatic mixed precision (AMP) with float16 computation. All experiments are conducted on $2{\times}$ Tesla T4 GPUs in the Kaggle environment.

The student backbone is initialized from the pretrained DeiT-S checkpoint, while its classification head is initialized using the trained full-token baseline classification head. The saliency-head linear layers are initialized from a truncated normal distribution with standard deviation $0.02$, with biases initialized to zero. During epoch 1, the backbone is frozen, and only the saliency heads and classification head are optimized; from epoch 2 onward, all parameters are fine-tuned.

\section{Results}\label{sec:results}

\subsection{Main Accuracy Results}

Table~\ref{tab:per_seed} presents per-seed top-1 accuracy of the baseline and SWAM, while Table \ref{tab:agg} summarizes the corresponding mean differences across seeds. Under the conservative schedule SWAM matches or surpasses the baseline across all three datasets and all seeds.  The most consistent improvement is on ImageNet-100 ($+0.26\%$ mean).  Under the aggressive schedule results are more evident. SWAM continues to gain on ImageNet-100 ($+0.23\%$ mean across three seeds) but incurs small losses on CIFAR-100 and TinyImageNet. These results indicate that the conservative schedule provides a more stable accuracy-efficiency tradeoff.

\begin{table}[!t]
  \centering
  \caption{Per-seed accuracy results.
  $\Delta$ = SWAM top-1 $-$ Baseline top-1 (percentage points). C: Conservative A: Aggressive}
  \label{tab:per_seed}
  
  \resizebox{\columnwidth}{!}{%
  \begin{tabular}{llccccc}
    \toprule
    Dataset & Sched. & Seed & Base. & SWAM & $\Delta$ & Speedup \\
    \midrule
    CIFAR-100     & C & 42 & 87.65\% & 87.69\% & $+0.04$ & $1.36{\times}$ \\
    CIFAR-100     & C & 43 & 87.56\% & 87.55\% & $-0.01$ & $1.36{\times}$ \\
    TinyImageNet & C & 42 & 88.25\% & 88.34\% & $+0.09$ & $1.35{\times}$ \\
    TinyImageNet & C & 43 & 88.08\% & 88.19\% & $+0.11$ & $1.37{\times}$ \\
    ImageNet-100  & C & 42 & 88.48\% & 88.88\% & $+0.40$ & $1.35{\times}$ \\
    ImageNet-100  & C & 43 & 88.74\% & 88.86\% & $+0.12$ & $1.35{\times}$ \\
    \midrule
    CIFAR-100     & A & 42 & $\approx87.5\%$ & $\approx87.4\%$ & $-0.13$ & $1.46{\times}$ \\
    TinyImageNet & A & 42 & 89.04\% & 88.86\% & $-0.18$ & $1.46{\times}$ \\
    TinyImageNet & A & 43 & 88.02\% & 87.95\% & $-0.07$ & $1.45{\times}$ \\
    ImageNet-100  & A & 42 & 88.56\% & 88.78\% & $+0.22$ & $1.45{\times}$ \\
    ImageNet-100  & A & 43 & 88.84\% & 88.92\% & $+0.08$ & $1.47{\times}$ \\
    ImageNet-100  & A & 44 & 88.72\% & 89.12\% & $+0.40$ & $1.44{\times}$ \\
    \bottomrule
  \end{tabular}%
  }
\end{table}

\begin{table}[!t]
  \centering
  \caption{Aggregated paired accuracy differences
  (mean and sample standard deviation in percentage
  points).}
  \label{tab:agg}
  \resizebox{\columnwidth}{!}{%
  \begin{tabular}{llcccc}
    \toprule
    Dataset & Schedule & Mean $\Delta$ & SD & Seeds & Range \\
    \midrule
    CIFAR-100 & Conservative & $+0.02$ & $0.04$ & 2 & $[-0.01,+0.04]$ \\
    TinyImageNet & Conservative & $+0.10$ & $0.01$ & 2 & $[+0.09,+0.11]$ \\
    ImageNet-100 & Conservative & $+0.26$ & $0.20$ & 2 & $[+0.12,+0.40]$ \\
    \midrule
    CIFAR-100 & Aggressive & $-0.13$ & -- & 1 & Single seed \\
    TinyImageNet & Aggressive & $-0.13$ & $0.08$ & 2 & $[-0.18,-0.07]$ \\
    ImageNet-100 & Aggressive & $+0.23$ & $0.16$ & 3 & $[+0.08,+0.40]$ \\
    \bottomrule
  \end{tabular}}
\end{table}

\subsection{Pair Scoring Strategy and Inference Efficiency}

Table~\ref{tab:ablation} presents results from similarity-only matching, additive saliency-aware scoring, and the proposed multiplicative max-saliency scoring, baselines and the proposed approach, respectively. Our proposed methods show improvement over the baselines. On CIFAR-100, the proposed multiplicative max-saliency scoring yields 87.69\% top-1 accuracy, whereas additive saliency-aware scoring and similarity-only matching score approximately 87.50\% and 87.25\%, respectively. The progression ToMe $\to$ SAM $\to$ SWAM shows consistent improvement: any saliency-based scoring outperforms pure similarity matching, and the multiplicative max formulation provides substantially stronger protection than additive mean scoring.  Table \ref{tab:efficiency} summarizes the inference efficiency of the baseline
and SWAM compression settings in terms of throughput, latency, FLOPs, and peak
GPU memory. Conservative SWAM achieves a $1.35{\times}$ throughput gain with 27.5\% FLOPs reduction.  Memory savings are modest ($  200$\,MB) because most model parameters remain unchanged, while the additional saliency heads introduce a small fixed overhead.

\begin{table}[!t]
  \centering
  \caption{Comparison of pair-scoring strategies on CIFAR-100 under the conservative setting (seed 42).}
  \label{tab:ablation}
  \begin{tabular}{lcc}
    \toprule
    Configuration & Top-1 & $\Delta$ vs Baseline \\
    \midrule
    Baseline DeiT-S (no reduction) & 87.65\% & --- \\
    ToMe (similarity only)         & $  87.25\%$ & $-0.40$ \\
    SAM (additive mean-saliency)   & $  87.50\%$ & $-0.15$ \\
    \textbf{SWAM (mult.\ max-sal.)} & \textbf{87.69\%} & $\mathbf{+0.04}$ \\
    \bottomrule
  \end{tabular}
\end{table}

\begin{table}[!t]
  \centering
  \caption{Inference efficiency (single GPU, batch 128).}
  \label{tab:efficiency}
  \begin{tabular}{lccccc}
    \toprule
    Config. & Throughput & Latency & FLOPs & Mem. & Speedup \\
    \midrule
    Baseline     & 282 img/s & 454\,ms & 4.25\,G & 3128\,MB & $1.00\times$ \\
    Conservative & 380 img/s & 336\,ms & 3.08\,G & 2926\,MB & $1.35\times$ \\
    Aggressive   & 412 img/s & 311\,ms & 2.82\,G & 2925\,MB & $1.46\times$ \\
    \bottomrule
  \end{tabular}
\end{table}

\section{Discussion and Limitations}\label{sec:discussion}

Our two proposed components play complementary roles in preserving informative token representations. Pair selection alone cannot prevent every informative token from being involved in a merge, particularly when the token reduction ratio becomes large. Asymmetric saliency-weighted merging provides an additional mechanism for preserving informative representations after a pair has been selected. This mechanism becomes more relevant under aggressive compression, where the reduction budget may require merging tokens that carry important information. SWAM performs more reliably under the conservative schedule, where sufficient less-informative tokens are available to meet the reduction budget. At more aggressive rates, the method relies increasingly on asymmetric aggregation and the quality of the learned saliency estimates.

\textbf{Limitations}: All experiments use DeiT-Small; generalization to DeiT-Tiny or DeiT-Base has not yet been evaluated. All experiments were conducted on the Kaggle platform due to hardware and storage quota constraints ($73$GB available vs.\ $150$GB required). SWAM EMA weights are initialized from the teacher's classification head, whereas the baseline EMA starts from random initialization.




\section{Conclusion}\label{sec:conclusion}

In this work, we introduced SWAM, a training-based token merging framework that incorporates multiplicative maximum-saliency pair ranking and asymmetric saliency-weighted aggregation. The two components provide complementary mechanisms for regulating token selection and preserving informative representations during progressive reduction. Lightweight saliency predictors introduce approximately $1\%$ parameter overhead, with explicit attention-rollout supervision applied only at the first reduction stage. Experiments on CIFAR-100, TinyImageNet, and ImageNet-100 demonstrate FLOPs reductions of $27.5$--$33.8\%$ and throughput speedups of $1.35$--$1.46\times$ under the evaluated compression schedules. The accuracy results show dataset- and
compression-dependent trade-offs. Future work will investigate generalization to the full ImageNet-1K dataset, larger and smaller ViT backbones, and alternative per-block reduction strategies.

\section*{Acknowledgments}
We gratefully acknowledge the Indian Institute of Information Technology Allahabad for providing the fellowship to pursue this research work.

\bibliographystyle{IEEEtran}
\bibliography{ref}

\begin{thebibliography}{10}
\providecommand{\url}[1]{#1}
\csname url@samestyle\endcsname
\providecommand{\newblock}{\relax}
\providecommand{\bibinfo}[2]{#2}
\providecommand{\BIBentrySTDinterwordspacing}{\spaceskip=0pt\relax}
\providecommand{\BIBentryALTinterwordstretchfactor}{4}
\providecommand{\BIBentryALTinterwordspacing}{\spaceskip=\fontdimen2\font plus
\BIBentryALTinterwordstretchfactor\fontdimen3\font minus \fontdimen4\font\relax}
\providecommand{\BIBforeignlanguage}[2]{{%
\expandafter\ifx\csname l@#1\endcsname\relax
\typeout{** WARNING: IEEEtran.bst: No hyphenation pattern has been}%
\typeout{** loaded for the language `#1'. Using the pattern for}%
\typeout{** the default language instead.}%
\else
\language=\csname l@#1\endcsname
\fi
#2}}
\providecommand{\BIBdecl}{\relax}
\BIBdecl

\bibitem{dosovitskiy2021vit}
A.~Dosovitskiy, L.~Beyer, A.~Kolesnikov, D.~Weissenborn, X.~Zhai, T.~Unterthiner, M.~Dehghani, M.~Minderer, G.~Heigold, S.~Gelly, J.~Uszkoreit, and N.~Houlsby, ``An image is worth 16x16 words: Transformers for image recognition at scale,'' in \emph{International Conference on Learning Representations}, 2021.

\bibitem{bolya2023tome}
D.~Bolya, C.-Y. Fu, X.~Dai, P.~Zhang, C.~Feichtenhofer, and J.~Hoffman, ``Token merging: Your vit but faster,'' in \emph{International Conference on Learning Representations (ICLR)}, 2023.

\bibitem{liang2022evit}
Y.~Liang, C.~Ge, Z.~Tong, Y.~Song, J.~Wang, and P.~Xie, ``Not all patches are what you need: Expediting vision transformers via token reorganizations,'' in \emph{International Conference on Learning Representations (ICLR)}, 2022.

\bibitem{wei2023tps}
S.~Wei, T.~Ye, S.~Zhang, Y.~Tang, and J.~Liang, ``Joint token pruning and squeezing towards more aggressive compression of vision transformers,'' in \emph{2023 IEEE/CVF Conference on Computer Vision and Pattern Recognition (CVPR)}.\hskip 1em plus 0.5em minus 0.4em\relax IEEE, 2023, pp. 2092--2101.

\bibitem{lee2024mctf}
S.~Lee, J.~Choi, and H.~J. Kim, ``Multi-criteria token fusion with one-step-ahead attention for efficient vision transformers,'' in \emph{2024 IEEE/CVF Conference on Computer Vision and Pattern Recognition (CVPR)}.\hskip 1em plus 0.5em minus 0.4em\relax IEEE, 2024, pp. 15\,741--15\,750.

\bibitem{abnar2020rollout}
S.~Abnar and W.~Zuidema, ``Quantifying attention flow in transformers,'' in \emph{Proceedings of the 58th annual meeting of the association for computational linguistics}, 2020, pp. 4190--4197.

\bibitem{touvron2021deit}
H.~Touvron, M.~Cord, M.~Douze, F.~Massa, A.~Sablayrolles, and H.~J{\'e}gou, ``Training data-efficient image transformers \& distillation through attention,'' in \emph{International conference on machine learning}.\hskip 1em plus 0.5em minus 0.4em\relax PMLR, 2021, pp. 10\,347--10\,357.

\bibitem{rao2021dynamicvit}
Y.~Rao, W.~Zhao, B.~Liu, J.~Lu, J.~Zhou, and C.-J. Hsieh, ``Dynamicvit: Efficient vision transformers with dynamic token sparsification,'' \emph{Advances in neural information processing systems}, vol.~34, pp. 13\,937--13\,949, 2021.

\bibitem{yin2022avit}
H.~Yin, A.~Vahdat, J.~M. Alvarez, A.~Mallya, J.~Kautz, and P.~Molchanov, ``A-vit: Adaptive tokens for efficient vision transformer,'' in \emph{2022 IEEE/CVF Conference on Computer Vision and Pattern Recognition (CVPR)}.\hskip 1em plus 0.5em minus 0.4em\relax IEEE, 2022, pp. 10\,799--10\,808.

\bibitem{iared}
B.~Pan, R.~Panda, Y.~Jiang, Z.~Wang, R.~Feris, and A.~Oliva, ``Ia-red\textsuperscript{2}: Interpretability-aware redundancy reduction for vision transformers,'' \emph{Advances in neural information processing systems}, 2021.

\bibitem{xu2022evo}
Y.~Xu, Z.~Zhang, M.~Zhang, K.~Sheng, K.~Li, W.~Dong, L.~Zhang, C.~Xu, and X.~Sun, ``Evo-vit: Slow-fast token evolution for dynamic vision transformer,'' in \emph{Proceedings of the AAAI conference on artificial intelligence}, vol.~36, no.~3, 2022, pp. 2964--2972.

\bibitem{ats}
M.~Fayyaz, S.~A. Koohpayegani, F.~R. Jafari, S.~Sengupta, H.~R.~V. Joze, E.~Sommerlade, H.~Pirsiavash, and J.~Gall, ``Adaptive token sampling for efficient vision transformers,'' in \emph{European conference on computer vision}, 2022.

\bibitem{spvit}
Z.~Kong, P.~Dong, X.~Ma, X.~Meng, W.~Niu, M.~Sun, X.~Shen, G.~Yuan, B.~Ren, H.~Tang \emph{et~al.}, ``Spvit: Enabling faster vision transformers via latency-aware soft token pruning,'' in \emph{European conference on computer vision}, 2022.

\bibitem{patch_slimming}
Y.~Tang, K.~Han, Y.~Wang, C.~Xu, J.~Guo, C.~Xu, and D.~Tao, ``Patch slimming for efficient vision transformers,'' in \emph{Proceedings of the IEEE/CVF Conference on Computer Vision and Pattern Recognition (CVPR)}, June 2022, pp. 12\,165--12\,174.

\bibitem{ryoo2021tokenlearner}
M.~Ryoo, A.~Piergiovanni, A.~Arnab, M.~Dehghani, and A.~Angelova, ``Tokenlearner: Adaptive space-time tokenization for videos,'' \emph{Advances in neural information processing systems}, vol.~34, pp. 12\,786--12\,797, 2021.

\bibitem{deepcompress}
S.~Ahmed, A.~Al~Arafat, D.~Najafi, A.~Mahmood, M.~N. Rizve, M.~Al~Nahian, R.~Zhou, S.~Angizi, and A.~S. Rakin, ``Deepcompress-vit: Rethinking model compression to enhance efficiency of vision transformers at the edge.''

\bibitem{tinynext}
F.~Zeng, H.~Li, J.~Guan, R.~Fan, T.~Wu, X.~Wang, and R.~Lai, ``An efficient hybrid vision transformer for tinyml applications,'' in \emph{Proceedings of the IEEE/CVF International Conference on Computer Vision}, 2025, pp. 19\,914--19\,924.

\bibitem{b16}
X.~Xu, S.~Wang, Y.~Chen, Y.~Zheng, Z.~Wei, and J.~Liu, ``Gtp-vit: Efficient vision transformers via graph-based token propagation,'' in \emph{Proceedings of the IEEE/CVF Winter Conference on Applications of Computer Vision}, 2024.

\bibitem{b17}
S.~Lee, J.~Choi, and H.~J. Kim, ``Multi-criteria token fusion with one-step-ahead attention for efficient vision transformers,'' in \emph{Proceedings of the IEEE/CVF conference on computer vision and pattern recognition}, 2024.

\bibitem{bonnaerens2023ltmp}
M.~Bonnaerens and J.~Dambre, ``Learned thresholds token merging and pruning for vision transformers,'' in \emph{Transactions on Machine Learning Research}, 2023.

\end{thebibliography}

\end{document}